\documentclass[11pt,a4paper]{article}
\usepackage[hyperref]{ranlp2025}
\usepackage{times}
\usepackage{latexsym}

\usepackage{microtype}
\usepackage{CJKutf8}
\usepackage{amsmath}
\usepackage{svg}

\usepackage{tikz}

\usepackage{enumitem}

\usepackage{textcomp}  % Required for encoding \textbigcircle
\usepackage{scalerel}  % Required for emoji \scalerel
\def\silveremoji{\scalerel*{\includegraphics{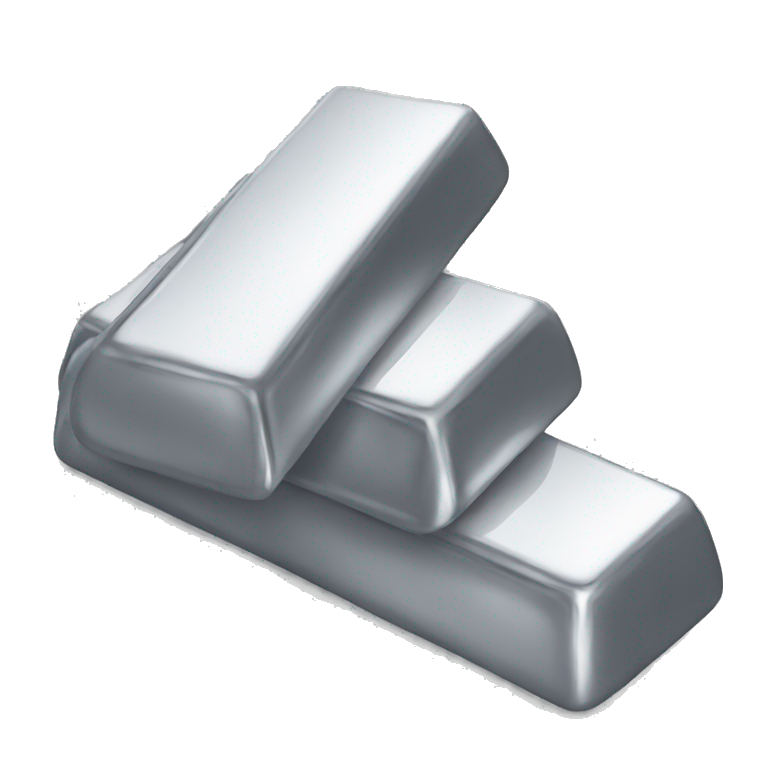}}{\textrm{\textbigcircle}}}
\NewDocumentCommand\scorename{}{\silveremoji \textsc{SiLVERScore}}

\aclfinalcopy % Uncomment this line for the final submission
\def\aclpaperid{224} %  Enter the acl Paper ID here

\title{\silveremoji~SiLVERScore: Semantically-Aware Embeddings \\ for Sign Language Generation Evaluation}

\author{
    Saki Imai \qquad Mert İnan \qquad Anthony Sicilia \qquad Malihe Alikhani \\
    Northeastern University, Boston MA \\
    \texttt{\{imai.s, m.alikhani\}@northeastern.edu}
}

\date{}

\begin{document}
\maketitle
\begin{abstract}
Evaluating sign language generation is often done through back-translation, where generated signs are first recognized back to text and then compared to a reference using text-based metrics. However, this two-step evaluation pipeline introduces ambiguity: it not only fails to capture the multimodal nature of sign language—such as facial expressions, spatial grammar, and prosody—but also makes it hard to pinpoint whether evaluation errors come from sign generation model or the translation system used to assess it. In this work, we propose \scorename, a novel semantically-aware embedding-based evaluation metric that assesses sign language generation in a joint embedding space. Our contributions include: (1) identifying limitations of existing metrics, (2) introducing SiLVERScore for semantically-aware evaluation, (3) demonstrating its robustness to semantic and prosodic variations, and (4) exploring generalization challenges across datasets. On PHOENIX-14T and CSL-Daily datasets, SiLVERScore achieves near-perfect discrimination between correct and random pairs (ROC AUC = 0.99, overlap $<$ 7\%), substantially outperforming traditional metrics\footnote{\url{https://github.com/sakimai/silverscore}}.
\end{abstract}

\section{Introduction}
\begin{figure}[t]
    \centering
    \includegraphics[width=\columnwidth]{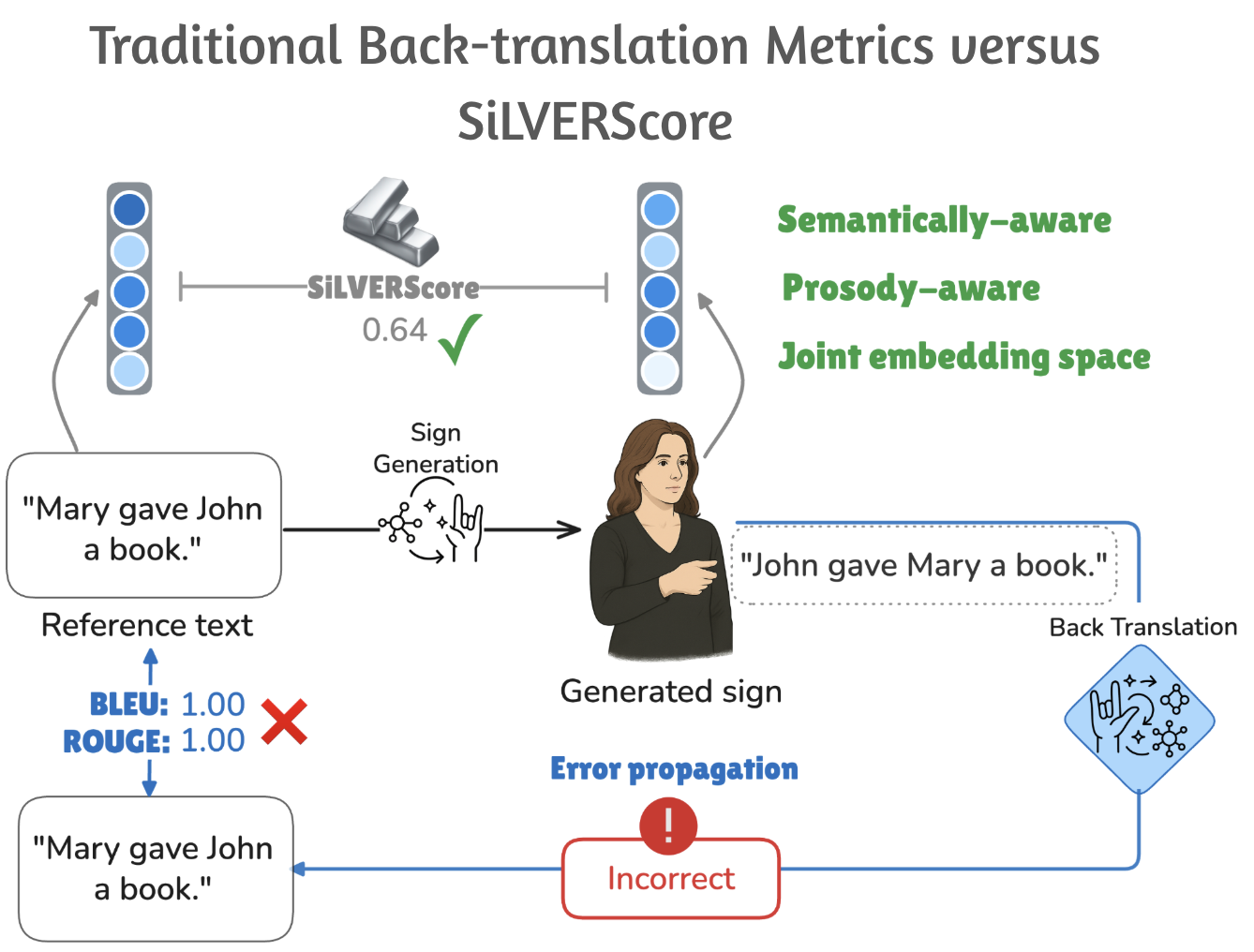}
    \caption{ In this example, a sign language generation model accidentally swaps the referents, generating the sign for “John gave Mary a book” instead of “Mary gave John a book.” Traditional text-based metrics like BLEU and ROUGE (bottom left) rely on back-translation and fail to catch the error, assigning a perfect score because the English output matches the reference text, even though the meaning is incorrect. In contrast, \scorename\ (top) compares the generated signing video directly with the reference using a joint embedding space, correctly identifying the error. }
    \label{fig:fig1}
\end{figure}

The ability to automatically evaluate sign language generation is critical for advancing accessibility and inclusion for the Deaf and Hard-of-Hearing (DHH) community, where collecting large scale human judgments remains expensive and challenging \cite{bragg2019sign, yin-etal-2021-including, huenerfauth2008evaluation}. Scalable and reliable evaluation is necessary to ensure that generated sign language content meets user needs. Yet, progress toward fully automated systems is hindered by the absence of effective evaluation methods \citep{liu2023survey}. To ensure outputs align with human expectations, we need robust evaluation metrics explicitly designed for the linguistic nature of sign language.

Automatically evaluating generated sign language remains challenging due to its unique multimodal linguistic nature, which incorporates facial expressions, manual markers, and spatiotemporal relationships into its prosody, iconicity, semantics, and pragmatics \cite{sandler2012phonological, Liddell_2003, huenerfauth2008evaluation}. Current evaluation methods rely on back-translation from visual to textual representations, which misaligns with the visual nature of sign language and leads to inaccuracies. As illustrated in Figure~\ref{fig:fig1}, a back‑translation metric can assign a perfect BLEU or ROUGE score even when the directional verb motion in the signing reverses the intended meaning.  This mismatch motivates an evaluation paradigm that inspects the video itself rather than its textual translation.

While embedding-based metrics such as BLEURT \cite{sellam2020bleurt}, BERTScore \cite{bert-score} and CLIPScore \cite{hessel-etal-2021-clipscore}, have shown success in natural language processing, they have been underexplored for sign language evaluation. This limitation is primarily due to the scarcity and domain specificity of sign language datasets, which restrict the generalizability of sign embeddings. We hypothesize that these data limitations have hindered the development of effective embedding-based metrics for sign language generation. 

To address this gap, we introduce \scorename\ (\textbf{Si}gn \textbf{L}anguage \textbf{V}ideo \textbf{E}mbedding \textbf{R}epresentation \textbf{Score}), a novel embedding-based metric for evaluating sign language generation. SiLVERScore directly compares generated and reference signs within a joint embedding space, capturing semantic and prosodic features.
 
Rather than asking whether embedding-based metrics are simply better than back-translation, our work investigates: \textit{how embedding-based evaluation can more faithfully capture the linguistic and prosodic nuances of sign language,} and under what conditions it offers robust and generalizable performance. Our contributions are as follows:
\begin{enumerate}[noitemsep, nolistsep]
    \item We survey existing evaluation metrics for sign language generation and highlight their limitations (\S~\ref{sec:related_work}).
    \item We introduce SiLVERScore, a novel semantically-aware embedding-based metric for evaluating sign language generation in a joint embedding space (\S~\ref{sec:signcicoscore}). 
    \item We conduct prosodic and semantic tests to demonstrate that SiLVERScore outperforms traditional metrics (\S~\ref{sec:semantic}, \S~\ref{sec:prosodic}).
    \item We perform a case study on generalization, the challenges of applying sign language models across different datasets and domains (\S~\ref{sec:generalization_problem}).
\end{enumerate}
\section{Survey of Evaluation Metrics for Sign Language Processing}
\label{sec:related_work}
The evaluation of sign language generation systems has traditionally relied on back-translation approaches, first introduced by \citet{Camgoz_2018_CVPR}. In these methods, a sign language translation model (typically trained by the authors) is used to convert the generated signs into text for evaluation. However, the absence of a standardized sign-to-text translation system complicates this approach, introducing unknown error propagation. 

To address these issues, researchers have proposed several multimodal metrics. For instance, Dynamic Time Warping Mean Joint Error \citep{huang2021towards} aligns generated and ground truth poses to measure spatial-temporal accuracy and compute the mean joint error. While effective for motion similarity, it penalizes valid linguistic variations that differ in pose but maintain semantic meaning. Similarly, Fréchet Gesture Distance \citep{Yoon2020SpeechGG}, Fréchet Video Distance \cite{unterthiner2019fvd}, Fréchet Inception Distance \cite{Heusel} compare gesture distributions but focus on physical similarity rather than semantics \citep{hwang2022non, 10483712, 10581980, 10581934}. Common video quality scores (SSIM, PSNR, Inception Score, Temporal Consistency Metric) measure image quality, diversity, or smoothness, ignoring whether the sign is linguistically correct \cite{natarajan2022development}.

In a visual-spatial SignWriting domain, signwriting-evaluation \cite{moryossef2024signwriting} was proposed as a metric designed for this by using its symbol distance metric using the Hungarian algorithm \cite{kuhn1955hungarian}. A sign language translation metric, SignBLEU \cite{kim2024signbleu} aims to mitigate the significant information loss due to the simplification to a single sequence of text for evaluation. However, despite its improvements, both remain confined to the text-realm. 

Multimodal embedding-based methods are promising due to their ability to capture multimodal elements and eliminate errors introduced by back-translation. Existing sign language embeddings, such as SignCLIP \cite{jiang-etal-2024-signclip}, offer a foundation for embedding-based evaluation. However, their application to sign language generation evaluation remains limited, primarily due to challenges in generalizability (\S~\ref{sec:generalization_problem}). This paper aims to bridge this gap by introducing and validating a semantically aware embedding-based evaluation metric tailored to sign language generation.
\section{\scorename}
\label{sec:signcicoscore}
The objective of SiLVERScore is to evaluate generated sign language videos without requiring a reference video. This evaluation measures the alignment between a sign video and its corresponding text by comparing their similarity in a shared joint embedding space, trained to capture multimodal relationships. The similarities are computed using CiCo \cite{Cheng}, a model that leverages contrastive learning to align video and text representations.
This approach addresses the alignment issues discussed in \S~\ref{sec:generalization_problem} by using a sliding window mechanism to localize alignment between modalities.

We employ CiCo due to its framework that: (i) formulates sign language retrieval as a cross-lingual retrieval task; (ii) demonstrates state-of-the-art performance on benchmarks such as PHOENIX-14T, CSL-Daily, and How2Sign; (iii) avoids reliance on pose estimation tools, eliminating dependency on pose extraction quality; and (iv) provides accessible code for implementation. 

\paragraph{Model Details}
The sign encoder processes sign videos using a sliding window mechanism to generate embeddings. This approach enables the model to handle continuous video streams without requiring explicit segmentation at test time. This encoder combines domain-agnostic features, captured by a pre-trained I3D network \cite{Varol} on BSL-1K, with domain-aware features from the same network fine-tuned on PHOENIX-14T/CSL-Daily. The features are weighted and fused before being fed into a 12-layer Transformer initialized with CLIP’s ViT-B encoder. The corresponding text is lowercased, byte pair encoded, and translated into English using Google Translate to align with the CLIP pretraining. The video and text embeddings are aligned through a contrastive learning objective with the InfoNCE loss. CiCo aligns video and text embeddings through a contrastive learning objective based on InfoNCE loss, which maximizes the similarity of matched video-text pairs while minimizing the similarity of unmatched pairs. This alignment is performed both globally across entire videos and texts and locally by retaining fine-grained mappings between video segments and individual text tokens. The resulting embeddings represent a semantically and temporally aware shared space that effectively captures the relationships between sign videos and their corresponding text annotations. 

\paragraph{Global Similarity Calculation}
Global similarity is derived from a fine-grained similarity matrix $E \in R^{M \times L}$:
\begin{equation}
E(i,j) = S_i \cdot W_j^T,
\end{equation}
where $S_i \in R^D$ and $W_j \in R^D$ represent video clip and word embeddings, respectively. To emphasize similarities, softmax re-weighting is applied:
\begin{equation}
E'(i,j) = \text{Softmax}(E(i,j)) \cdot E(i,j).
\end{equation}
Row-wise summation followed by averaging yields the video-to-text similarity $Z_{V2T}$, while column-wise operations yield the text-to-video similarity $Z_{T2V}$. 

In the implementation, the $Z_{V2T}$ and $Z_{T2V}$ similarities are equally weighted in the loss function. This equal weighting ensures that the global alignment of video-to-text and text-to-video pairs is equivalent, making it sufficient to use either $Z_{V2T}$ or $Z_{T2V}$ as the similarity metric. Without loss of generality, we use $Z_{V2T}$ for our similarity metric.

\paragraph{Scaling for Interpretability}
To ensure comparability with metrics like BLEU and ROUGE, we follow a similar approach to CLIP-Score by scaling the embeddings with a weighting factor of 3.5, expanding the score distribution range to [0,100].
\section{Experiments}
To evaluate the effectiveness of SiLVERScore, we conduct multiple experiments to assess the performance compared to back-translation methods.

\paragraph{Datasets}
1) PHOENIX-14T dataset \cite{Camgoz_2018_CVPR} is widely recognized as the benchmark dataset for sign language generation \cite{progressive_t, saunders2, viegas-etal-2023-including, inan-etal-2022-modeling}.  It consists of German Sign Language weather forecast videos segmented into sentences, accompanied by corresponding German transcripts and sign-gloss annotations.
2) CSL-Daily \cite{zhou2021improving}. To broaden the domain beyond weather forecasts, we include CSL-Daily, a dataset covering Chinese Sign Language in various daily-life scenarios. This enables us to test the generalizability of SiLVERScore to diverse real-world contexts.

\paragraph{Translation Model}
For the back translation model, we use the multi-stream keypoint attention network proposed by \citealp{Guan2024MultiStreamKA}, due to its state-of-the-art performance in sign language translation. This approach minimizes the error propagation caused by inaccuracies in back translation.

\paragraph{Metrics}
We evaluate the quality of back-translated text using both rule-based and embedding-based metrics. For rule-based evaluation, we compute BLEU scores with sacreBLEU \cite{post-2018-call} and ROUGE scores. For embedding-based evaluation, we use BLEURT (specifically BLEURT-20, \citealp{pu2021learning}) and BERTScore (using the \texttt{bert-base-multilingual-cased} model to accommodate the German and Chinese dataset; \citet{bert-score}). These metrics provide a benchmark for assessing the alignment quality of SiLVERScore in comparison to traditional back-translation evaluation methods.

\subsection{Which metric can distinguish between correct and random video-text pairs?} \label{sec:distr}
\subsubsection{Distribution of Metric Scores}
\begin{figure*}[ht]
    \centering
    \includegraphics[width=0.325\linewidth]{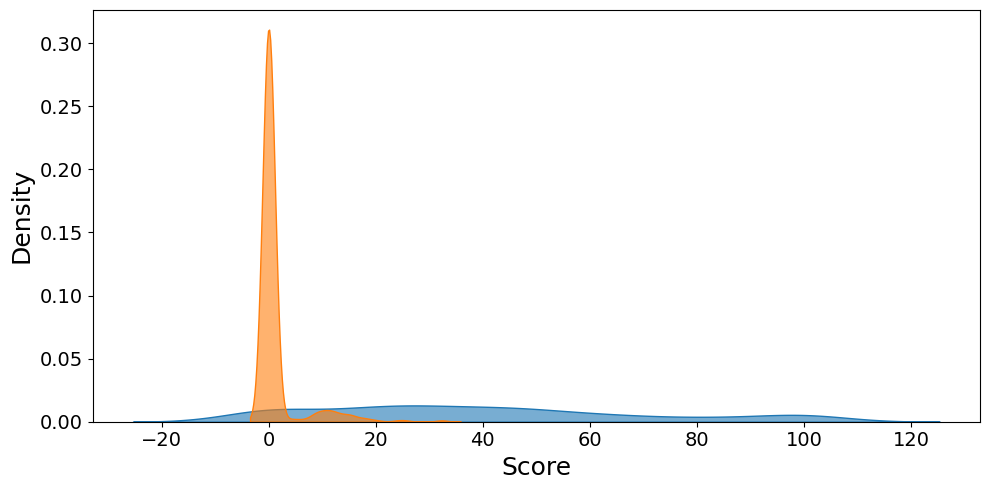}
    \includegraphics[width=0.325\linewidth]{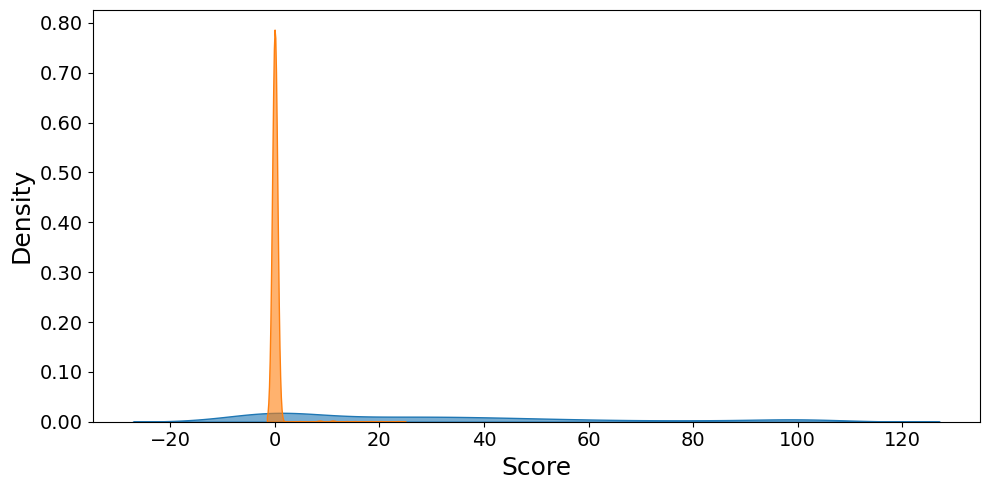}
    \includegraphics[width=0.325\linewidth]{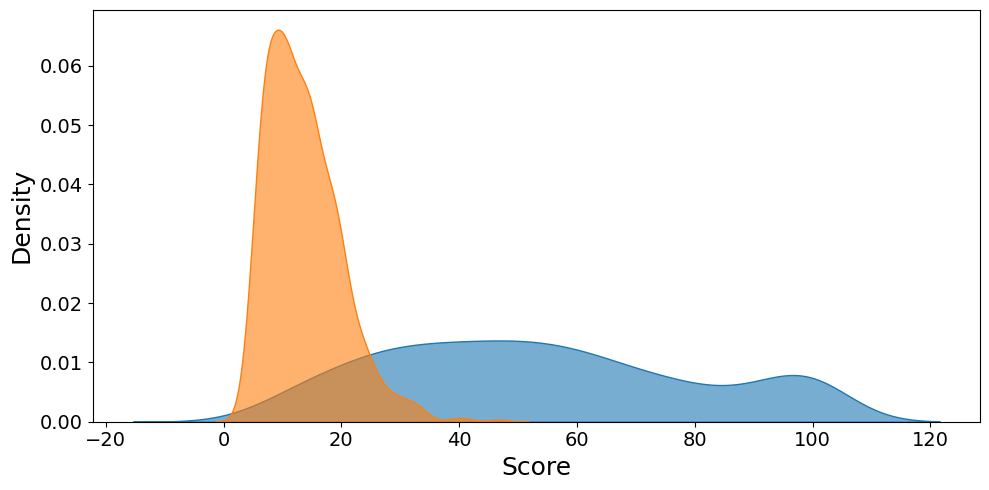}
    
    \includegraphics[width=0.325\linewidth]{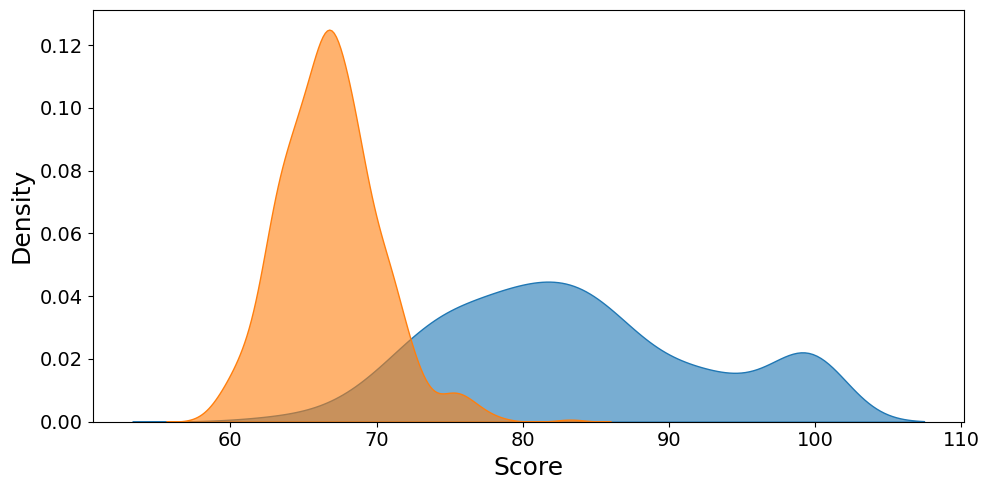}
    \includegraphics[width=0.325\linewidth]{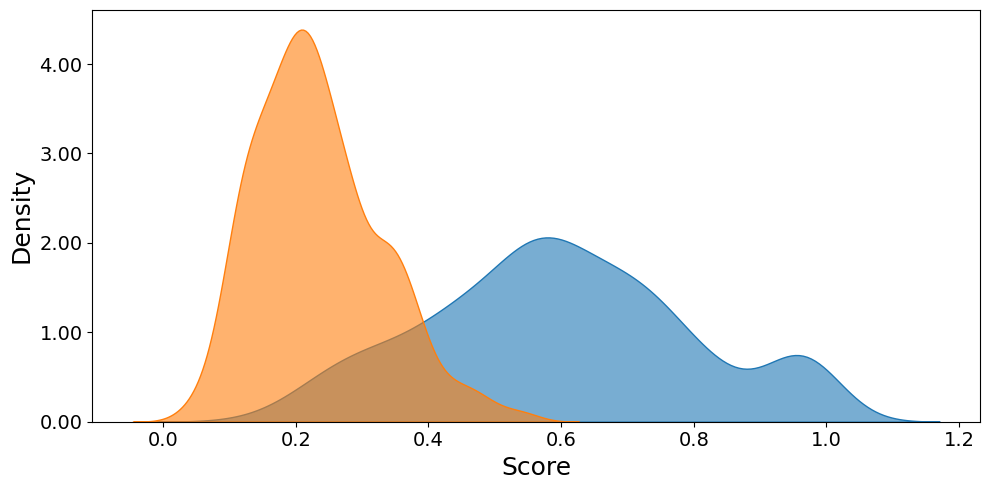}
    \includegraphics[width=0.325\linewidth]{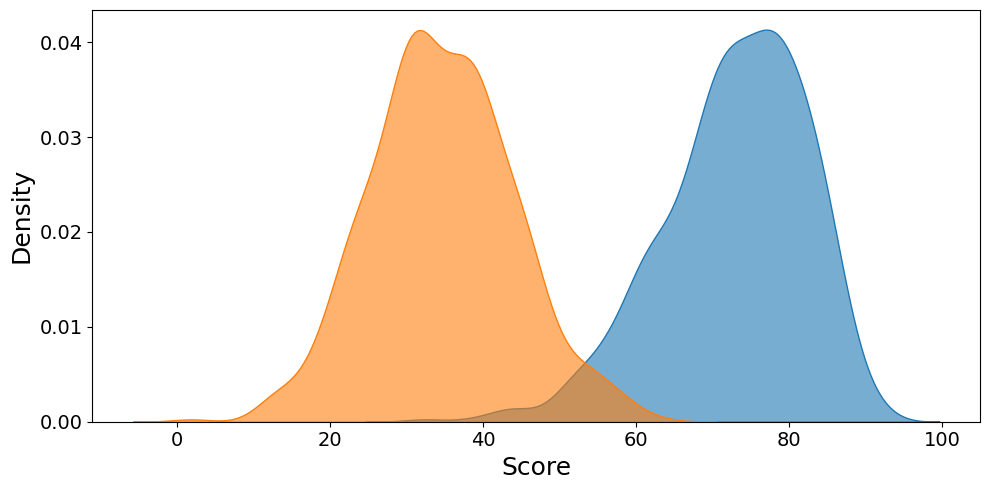}
    
    \caption{Kernel Density Plots for different metrics. \textbf{Top row (left to right, rule-based metrics):} BLEU-2, BLEU-3, ROUGE. \textbf{Bottom row (left to right, embedding-based metrics):} BERTScore, BLEURT, SiLVERScore. The blue curve represents the distribution of scores for matching indices (aligned pairs), while the orange curve represents different indices (misaligned pairs). SiLVERScore exhibits a clear separation between the two distributions, indicating a strong ability to distinguish aligned from misaligned pairs. In contrast, BLEU and ROUGE metrics show more overlap, reflecting their sensitivity to surface-level variations.}
    \label{fig:kdp}
\end{figure*}

To qualitatively evaluate the performance of different metrics, we analyze the kernel density plots in Figure~\ref{fig:kdp}. These plots illustrate the distribution of scores for correctly matched video-text pairs (\textcolor{blue}{blue curve}) and randomly paired samples (\textcolor{orange}{orange curve}).
SiLVERScore shows a clear separation between the two distributions, with minimal overlap. This indicates its strong ability to distinguish aligned pairs from misaligned ones. In contrast, BLEU-2 exhibits significant overlap, particularly for lower score ranges, suggesting reduced discriminative power for this task. Similarly, the ROUGE shows partial separation but retains overlap between the two distributions. BERTScore and BLEURT show improved separation compared to rule-based metrics but still exhibit some overlap. The sharp distinction and density clustering of scores in the SiLVERScore plot indicate its effectiveness in capturing semantic alignment between video and text representations. Figure~\ref{fig:kdp} focuses on PHOENIX-14T, but we observe similar trends on CSL-Daily. (Additional density plots for this condition and the rest of the metrics are in Appendix~\ref{sec:density-plots}.)

\subsubsection{Quantifying overlap and separability}
To complement the qualitative insights from the kernel density plots, we quantify the ability of each metric to distinguish between correctly aligned and randomly paired samples using overlap percentage and ROC AUC (Receiver Operating Characteristic Area Under the Curve). The results are summarized in Table~\ref{tab:merged_metric_comparison}. BLEU-4 is omitted for CSL-Daily because consecutive 4-character n-grams in Chinese can lead to sparse counts, producing \texttt{NaN} in the calculation. Since each metric operates on a different scale, we applied Min-Max normalization to scale all metrics to the [0,1] range for a fair comparison.

\begin{table*}[h]
    \centering
    \resizebox{\textwidth}{!}{ 
    \begin{tabular}{lcccccccc}
        \hline
        & \multicolumn{4}{c}{\textbf{Correct vs. Random (\S~\ref{sec:distr})}} & \multicolumn{4}{c}{\textbf{Original vs. Reordered (\S~\ref{sec:semantic})}} \\
        & \multicolumn{2}{c}{PHOENIX-14T} & \multicolumn{2}{c}{CSL-Daily} & \multicolumn{2}{c}{PHOENIX-14T} & \multicolumn{2}{c}{CSL-Daily} \\
        \cline{2-9}
        & Overlap \textdownarrow & AUC \textuparrow & Overlap \textdownarrow & AUC \textuparrow & Overlap \textuparrow & AUC \textdownarrow & Overlap \textuparrow & AUC \textdownarrow \\
        \hline
        BLEU-1      & 19.78  & 0.95   & 6.04   & \textbf{0.99}  & 64.49  & 0.65   & 69.28  & 0.45 \\
        BLEU-2      & 24.30  & 0.90   & 5.27   & 0.98  & 71.50  & 0.63   & 69.60  & 0.49 \\
        BLEU-3      & 38.63  & 0.81   & 23.89  & 0.88  & 66.98  & 0.65   & 76.43  & 0.52 \\
        BLEU-4      & 55.45  & 0.72   & -      & -     & 69.47  & 0.63   & 83.91  & 0.53 \\
        ROUGE       & 19.94  & 0.95   & 6.12   & 0.99  & 67.45  & 0.67   & 70.05  & 0.54 \\
        BERTScore   & 14.17  & 0.97   & 9.27   & 0.98  & 78.19  & 0.55   & 75.18  & 0.51 \\
        BLEURT      & 21.65  & 0.95   & 11.90  & 0.98  & 81.31  & \textbf{0.47}   & 70.24  & \textbf{0.39} \\
        SiLVERScore & \textbf{6.85}   & \textbf{0.99}   & \textbf{7.40}   & \textbf{0.99}  & \textbf{83.49}  & 0.60   & \textbf{87.84}  & 0.45 \\
        \hline
    \end{tabular}
    } 
    \caption{Comparison of overlap percentages and ROC AUC for various metrics across PHOENIX-14T and CSL-Daily. In ``Correct vs.\ Random'' (left columns), lower Overlap and higher AUC reflect better discrimination, and SiLVERScore achieves minimal Overlap (6.85--7.40\%) and near-maximal AUC (0.99).  In ``Original vs.\ Reordered'' (right columns), higher Overlap and lower AUC indicate greater tolerance to meaning-preserving reorderings, where SiLVERScore also achieves the highest Overlap (83.49--87.84\%) and lower ROC AUC.}
    \label{tab:merged_metric_comparison}
\end{table*}

\paragraph{Overlap percentage}
Overlap percentage measures how much the distributions of scores for correct and random pairs intersect. Lower overlap percentages indicate better discriminative power. SiLVERScore achieves the lowest overlap 6.85\% on PHOENIX-14T, and 7.40\% on CSL-Daily. BLEU-1 and ROUGE also show single-digit overlaps on CSL-Daily, yet its kernel density plots show that these distributions remain widely dispersed. BERTScore and BLEURT remain competitive with low overlaps, but neither is consistently smaller than SiLVERScore.

\paragraph{ROC AUC}
ROC AUC measures the metric's ability to distinguish between the two distributions. Higher ROC AUC values indicate better separability, with a maximum value of 1.0.  SiLVERScore attains 0.99 AUC for correct vs. random pairs on both datasets, confirming the separation observed in density plots. Overall, the results show that learned embedding-based metrics (SiLVERScore, BERTScore, BLEURT) outperform rule-based metrics in distinguishing between correctly aligned and misaligned video-text pairs.

\subsection{Which metric captures semantic distinctions through targeted changes?}\label{sec:semantic}
Rule-based metrics are inherently sensitive to the exact ordering of words, even when the overall meaning remains unchanged. To demonstrate this sensitivity, we designed an experiment where GPT-4o \cite{hurst2024gpt} was used to reorder words in sentences while preserving their meaning. The exact prompt provided to GPT-4o was:
\vspace{-3pt}
\begin{quote}
\textit{Reorder the words in the following sentence while keeping the meaning the same:} \textit{\{text\}}
\textit{Reordered sentence:}
\end{quote}
\vspace{-10pt}

\begin{figure}[h]
    \centering
    \includegraphics[width=1\linewidth]{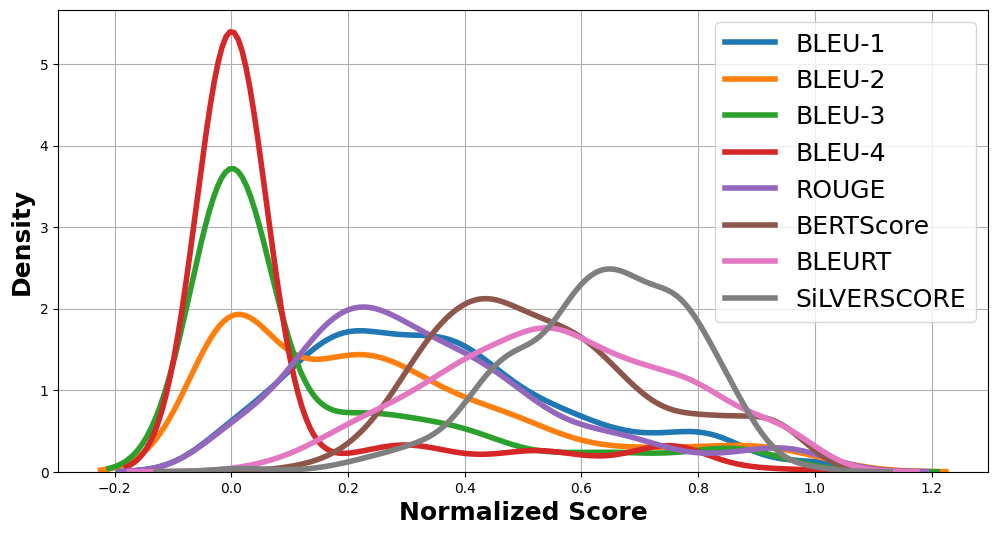}
    \caption{Kernel Density plots comparing the score distributions of different evaluation metrics when applied reordered hypotheses. SiLVERScore, BERTScore, and BLEURT show broader distributions and higher overlap, while rule-based metrics exhibit sharp peaks at lower scores. This indicates their sensitivity to surface-level word order changes.}
    \label{fig:kde_reordered}
\end{figure}

\paragraph{Kernel density plot}
Figure~\ref{fig:kde_reordered} illustrates how different metrics respond to surface-level changes, specifically word reordering, on PHOENIX-14T (plot for CSL-Daily in Appendix~\ref{sec:csl-daily-app}). SiLVERScore exhibits the highest score distribution, suggesting its robustness to reordering and its ability to capture semantic content. In contrast, BLEU and ROUGE display sharp peaks and narrower distributions concentrated in the lower score range. This pattern exhibits a clear distinction between rule-based and embedding-based metrics.

\paragraph{Quantifying overlap and separability}
In this experiment, the scores are computed by comparing the ground-truth references with their corresponding hypotheses. While these hypotheses may contain errors, they represent the best available approximations of the ground truth. Lower ROC AUC values indicate that the metric maintains its scores despite reordering, reflecting robustness to surface-level variations.

In Table~\ref{tab:merged_metric_comparison}, SiLVERScore demonstrates the highest overlap across both datasets (83.49\% on PHOENIX-14T, 87.84\% on CSL-Daily). This indicates its ability to recognize reordered sentences as semantically equivalent. In contrast, BLEU-1 and BLEU-2 exhibit much lower overlaps, confirming their strict reliance on word order rather than meaning. Moreover, SiLVERScore achieves relatively low ROC AUCs, suggesting it better maintains robustness to reordering.

It is important to note that the original distribution contains errors, which may affect the Overlap and ROC AUC values for all metrics. This could explain why SiLVERScore's ROC AUC is slightly higher than those of other metrics.

\begin{figure*}[h]
    \centering
    \includegraphics[width=1\linewidth]{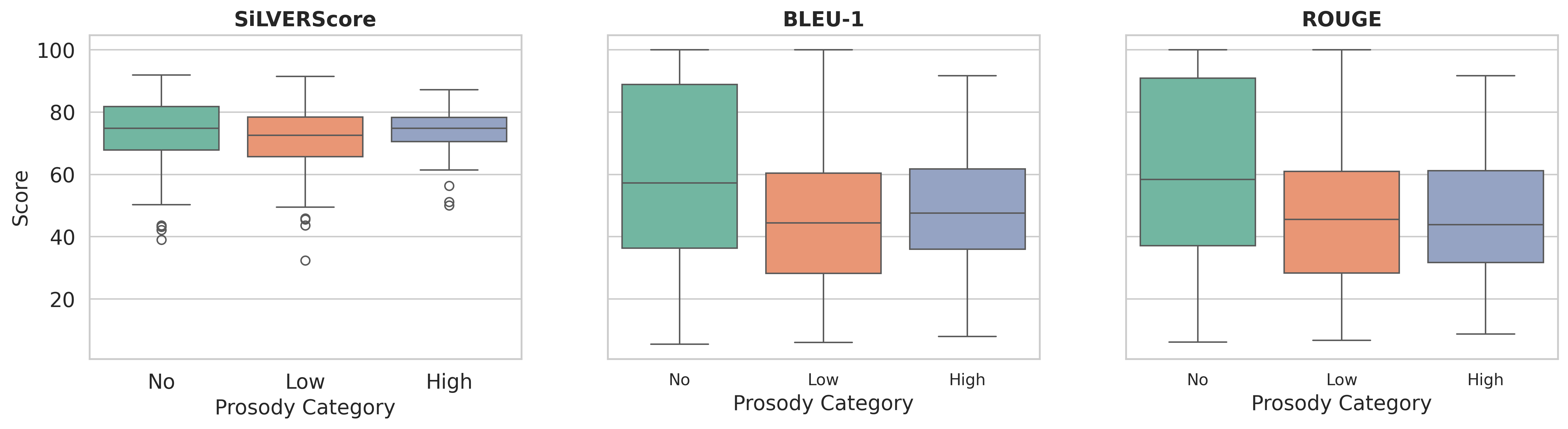}
    \caption{Box plots showing the distribution of SiLVERScore, BLEU-1, and ROUGE scores across three prosody intensity categories (No Intensity, Low Intensity, and High Intensity). While SiLVERScore remains stable across all categories, both BLEU-1 and ROUGE exhibit a noticeable decline in scores as prosody intensity increases. This drop suggests that BLEU-1 and ROUGE are sensitive to prosodically-rich sentences.}
    \label{fig:prosody_boxplots}
\end{figure*}

\subsection{Which metric can evaluate multimodal and pragmatic aspects more effectively?}
\label{sec:prosodic}
\subsubsection{Motivation and Setup} Sign languages rely heavily on prosodic markers such as facial expressions, pauses, and intensity to convey meaning. Evaluating the robustness of metrics to prosodic variations is critical, as traditional back-translation-based methods often fail to capture such multimodal cues. We build on the work of \citealp{inan-etal-2022-modeling}, which provided human-annotated token-level prosody intensities for the PHOENIX-14T dataset. These annotations classify tokens into three distinct prosodic levels: (i) no intensity: 0, indicating the absence of prosodic markers; (ii) low intensity: 1, reflecting a low degree of intensity markers; and (iii) high intensity: 2, representing high-degree intensity markers.

\paragraph{Sentence level prosody} We define sentence intensity as the sum of the intensity levels of its tokens, \( I = \sum_{i=1}^{n} t_i \), where \( t_i \) is the intensity of token \( i \). Sentences are categorized into three prosody levels: No Intensity \( I = 0 \), Low Intensity \( 1 \leq I \leq 4 \), and High Intensity \( I \geq 5 \).

\paragraph{Prosody level distribution} The dataset exhibits the following distribution of sentences across these prosody categories: 328 sentences (51.09\%) fall under No Intensity, 238 sentences (37.07\%) under Low Intensity, and 76 sentences (11.84\%) under High Intensity. This distribution indicates that the majority of sentences either lack prosodic markers or exhibit low levels of prosody, while highly expressive sentences are comparatively rare. 

\subsubsection{Distribution of Scores Across Prosody Categories}
To analyze the impact of prosody on evaluation metrics, we categorized sentences based on the sentence-level intensity sums defined earlier. Figure~\ref{fig:prosody_boxplots} shows the distributions of SiLVERScore, BLEU-1, and ROUGE scores across the categories. 

\paragraph{SiLVERScore Stability}
SiLVERScore remains consistent across the three prosody categories, showing minimal variation in median and interquartile range. This demonstrates that SiLVERScore effectively evaluates semantic alignment without being influenced by prosodic intensity.

\paragraph{BLEU-1 and ROUGE Sensitivity}
BLEU-1 and ROUGE scores decline with increasing prosody intensity, with median scores for High Intensity significantly lower than for No Intensity. This trend indicates that these metrics struggle with prosodically-rich sentences.

\paragraph{Score Variability}
Both BLEU-1 and ROUGE display higher variability in the High Intensity category, suggesting inconsistent performance in evaluating expressive signing.

\subsection{Correlation with Prosodic Intensity}
As shown in Table~\ref{tab:prosody_correlation_pvalue}, traditional back-translation-based metrics (BLEU and ROUGE) exhibit significant negative correlations with prosody intensity (e.g., BLEU-4: -0.200, p = $3.31 \times 10^{-7}$), reflecting their vulnerability to prosodic variations. This behavior reflects the limitations of traditional metrics, which depend on surface-level text alignment and are vulnerable to information loss during back translation. 

\begin{table}[!h]
\centering
\begin{tabular}{lcc}
\hline
\textbf{Metric}       & \textbf{Correlation} & \textbf{p-value}      \\
\hline
BLEU-1                & -0.160               & $<$ 0.01                \\
BLEU-2                & -0.178               & $<$ 0.01                \\
BLEU-3                & -0.191               & $<$ 0.01                \\
BLEU-4                & -0.200               & $<$ 0.01                \\
ROUGE                 & -0.179               & $<$ 0.01                \\
BERTScore             & -0.144               & $<$ 0.01                \\
BLEURT                & -0.101               & 0.01                \\
SiLVERScore             & -0.004               & 0.93                \\
\hline
\end{tabular}
\caption{Pearson Correlation and p-value of metrics with sentence-level prosody intensity.  SiLVERScore demonstrates no significant correlation while other metrics exhibit negative correlations with prosody intensity.}
\label{tab:prosody_correlation_pvalue}
\end{table}

In contrast, SiLVERScore exhibits no significant correlation with prosody intensity (correlation: -0.004, p = 0.9277), indicating its robustness to prosodic variations. This robustness suggests SiLVERScore’s ability to evaluate semantic alignment without being influenced by expressive elements.
\section{The Generalization Problem}
\label{sec:generalization_problem}
While evaluation metrics are expected to generalize across diverse datasets, this remains a significant challenge in sign language processing due to the limited size and diversity of available datasets. As highlighted by \citet{jiang-etal-2024-signclip}, one of the largest sign language datasets, SpreadtheSign, contains only 456,913 examples, which is orders of magnitude smaller than datasets in related domains (e.g., 400M examples for CLIP and 136M for VideoCLIP). In this section, we empirically demonstrate that even SignCLIP, the largest contrastive learning model to date, struggles with generalization at the token level.

\subsection{Evidence of Limited Generalization}
\begin{table*}[h]
    \centering
    \resizebox{.75\textwidth}{!}{%
    \begin{tabular}{l l c c c}
        \hline
        \textbf{Fine-tuned on} & \textbf{Tested on} & \textbf{R @ 1} & \textbf{R @ 5} & \textbf{R @ 10} \\
        \hline
        \multicolumn{5}{l}{\textbf{Token Level (\S~\ref{sec:token_level})}} \\
        - & Citizen & $1.40 \times 10^{-3}$ & $6.10 \times 10^{-3}$ & $1.12 \times 10^{-2}$ \\
        Citizen & Citizen & $6.39 \times 10^{-2}$ & $2.71 \times 10^{-1}$ & $4.39 \times 10^{-1}$ \\
        \multicolumn{5}{l}{\textbf{Sentence Level (\S~\ref{sec:sentence_level})}} \\
        WMTSLT & WMTSLT & $3.70 \times 10^{-3}$ & $1.75 \times 10^{-2}$ & $3.23 \times 10^{-2}$ \\
        \multicolumn{5}{l}{\textbf{Token Level Language Specific (\S~\ref{sec:lang_specific})}} \\
        Signs, SemLex & Citizen & $3.04 \times 10^{-5}$ & $5.00 \times 10^{-4}$ & $8.00 \times 10^{-4}$ \\
        Citizen, Signs, SemLex & Citizen & $4.36 \times 10^{-2}$ & $1.76 \times 10^{-1}$ & $2.88 \times 10^{-1}$ \\
        \multicolumn{5}{l}{\textbf{With SignCL (\S~\ref{sec:signcl})}} \\
        Signs, SemLex & Citizen & $9.11 \times 10^{-5}$ & $5.00 \times 10^{-4}$ & $9.00 \times 10^{-4}$ \\
        \multicolumn{5}{l}{\textbf{With Data Augmentation (\S~\ref{sec:aug})}} \\
        Signs, SemLex & Citizen & $0.00 \times 10^{0}$ & $2.00 \times 10^{-4}$ & $6.00 \times 10^{-4}$ \\
        Signs, SemLex & Citizen & $6.07 \times 10^{-5}$ & $9.11 \times 10^{-5}$ & $3.00 \times 10^{-4}$ \\
        \hline
    \end{tabular}%
    }
    \caption{Text-to-Video Retrieval results and generalization across datasets. Results are shown for different fine-tuning datasets and test datasets.}
    \label{tab:generalization_results}
\end{table*}

\subsubsection{Token Level Generalization}
\label{sec:token_level}
We evaluated SignCLIP on ASL Citizen \cite{Desai} and ASL Signs \cite
{asl-signs}. The results show that SignCLIP's generalization capability is limited without fine-tuning. (Descriptions of these datasets can be found in Appendix~\ref{sec:dataset_details}.)

\begin{figure}[htbp]
    \centering
    \includegraphics[width=\linewidth]{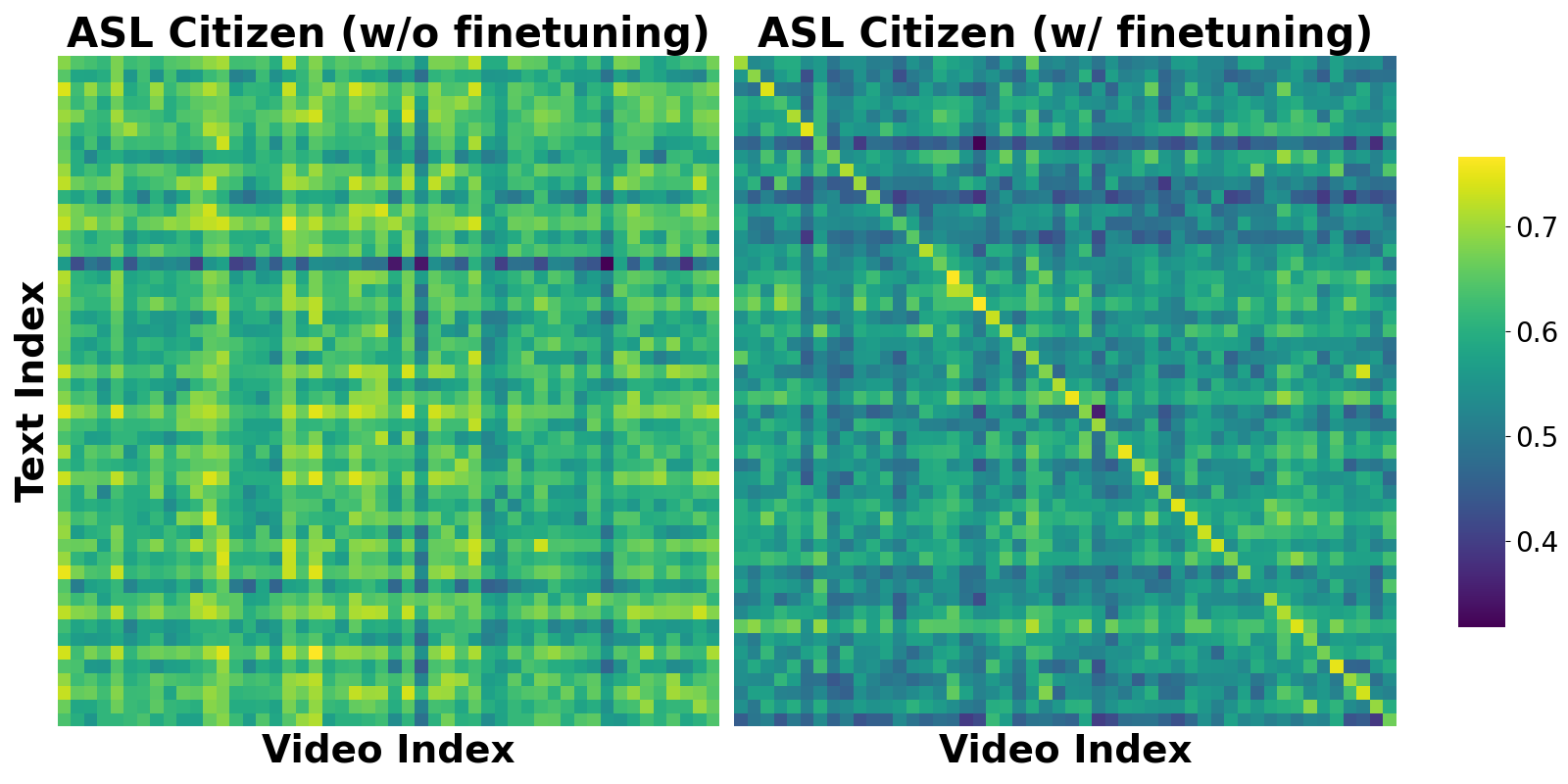}
    \caption{Heatmaps of SignCLIP embeddings cosine similarity scores for two datasets: ASL Citizen (token level) and WMTSLT (sentence level). \textbf{Left}: Finetuning increases alignment, as indicated by the clearer diagonal line. \textbf{Right}: After finetuning, the model appears to overfit, assigning high similarity scores to many pairs.}
    \label{fig:heatmaps}
\end{figure}

Figure~\ref{fig:heatmaps} illustrates the cosine similarity between video and text embeddings. Ideally, high similarity values should appear along the diagonal, indicating alignment between corresponding video-text pairs. Before fine-tuning, the heatmaps display low, diffuse similarity scores, indicating poor video-text alignment. Fine-tuning significantly improves alignment, indicating the necessity of dataset-specific adaptation. A similar trend is observed for ASL Signs (figures in Appendix~\ref{sec:heatmaps}).

\subsubsection{Sentence Level Generalization}
\label{sec:sentence_level}
We evaluated SignCLIP's sentence-level generalization on the WMTSLT Focus News Corpus \citep{muller_mathias_2022_6621480}. 
Despite fine-tuning, SignCLIP struggles to achieve strong results (R@1 = 0.0436). Heatmaps (in Appendix~\ref{sec:heatmaps}) reveal diffuse patterns before fine-tuning and overfitting after, due to the dataset's limited size (9000 instances).

\subsubsection{Token Level Language Specific Generalization}
\label{sec:lang_specific}
To investigate the effect of data size on generalization, we fine-tuned SignCLIP using combined training samples from ASL Signs and SemLex datasets. Despite this, SignCLIP fails to generalize effectively to ASL Citizen (R@5 = 0.0005). Even when training on all three datasets, the test set performance on ASL Citizen did not improve significantly. This suggests that dataset-specific characteristics influence performance even when substantial training data is available.

\subsubsection{Representation Density}
\label{sec:signcl}
\citealp{ye2024improving} identified a representation density problem, where the semantic visual representations of different sign gestures tend to be closely clustered together, making them hard to distinguish. The proposed contrastive learning strategy, SignCL, encourages the learning of discriminative feature representations. However, applying SignCL to our data yielded limited improvement in retrieval results (R@1 = $9.11 \times 10^{-5}$), compared to (R@1 = $3.04 \times 10^{-5}$) with vanilla contrastive learning.

\subsubsection{Data Augmentation}
\label{sec:aug}
Data augmentation is a commonly employed technique to improve model generalization, especially in domains with limited data. To this end, we experimented with several data augmentation strategies including: spatial 2D augmentation, temporal augmentation, and Gaussian noise on keypoints \citep{jiang-etal-2024-signclip}. Results show negligible gains (R@1 = 0 with 2D-aug; $6.07 \times 10^{-5}$ with temporal augmentation), highlighting the limitations of conventional augmentation techniques in enhancing generalization. This suggests that limited dataset diversity and the complexity of visual sign representations cannot be fully addressed through conventional augmentation techniques alone.

\subsection{How SiLVERScore Addresses Generalization Challenges}

Our findings from the experiments suggest the idea that, given current constraints in data availability, tailoring metrics to specific datasets is necessary to create alignment between text and sign. 

We proposed a dataset-specific evaluation metric designed to leverage the strengths of embedding-based methods while addressing the constraints of current sign language datasets. By optimizing for specific domains and datasets, we can achieve more reliable evaluations and better alignment with the linguistic and multimodal nature of sign language. 

\section{Conclusion}
Through the introduction of SiLVERScore, we demonstrated the empirical strengths of embedding-based methods, including robustness to semantic variation, prosodic intensity, and a more holistic multimodal evaluation. Our results show that SiLVERScore can overcome limitations of traditional back-translation metrics.

SiLVERScore has the potential to reshape sign language evaluation standards by advancing accessibility for the DHH community and promoting inclusivity in language technologies. Its robustness and semantic sensitivity make it well-suited for broader challenges in multimodal NLP, such as cross-lingual evaluation and integration with video generation models. To support open research and encourage further advancements, we release the code for SiLVERScore’s analysis and computation.

Future efforts should integrate insights from computer graphics, such as improved modeling of spatial relationships and prosody in sign language, to further refine embedding-based methods. Incorporating richer multimodal features, including gesture dynamics and temporal coherence, could enhance the evaluation of expressive and context-dependent signing. Additionally, addressing the scarcity of diverse, large-scale datasets remains critical for improving model generalization.
\section*{Limitations}
While the proposed metric, SiLVERScore, demonstrates strong empirical performance, this work has several limitations. One limitation is the absence of human evaluation. Although SiLVERScore shows clear advantages over traditional methods using back translation, it remains crucial to validate its alignment with human judgments. Human evaluators could provide insight into whether the metric effectively captures the semantic and linguistic aspects of generated sign language. Addressing this limitation will be a focus of future work.

Our approach translates textual annotations into English to leverage CLIP embeddings, which can be problematic for low-resource languages that lack reliable English translations. Although sign language processing itself remains under-resourced, future work must explore multilingual or language-agnostic embedding strategies to ensure SiLVERScore’s applicability beyond English-based contexts.

We currently assess alignment at the sentence level, which overlooks discourse-level dependencies and references across multiple sentences \cite{tanzer-etal-2024-reconsidering}. Sign language frequently encodes meaning through extended contexts, so capturing long-range dependencies will be crucial for a truly comprehensive evaluation. Nevertheless, introducing a semantically-aware multimodal metric at the sentence level is a key step beyond n-gram approaches and addresses error propagation from back-translation.

Finally, while the results show that prosody does not degrade SiLVERScore's performance, this does not imply that the metric explicitly models prosody. Future research should investigate how to incorporate explicit prosodic modeling into evaluation metrics to better capture the expressive nuances of sign language.

\paragraph{Potential Risks} Adopting embedding-based metrics can inadvertently inherit biases, stereotypes, or inaccuracies from the underlying training data and models. If the pre-trained embeddings contain demographic, cultural, or linguistic biases, these may influence evaluations and potentially disadvantage certain signers or signing styles. Moreover, inaccuracies introduced at the text-annotation stage could propagate through the metric, reinforcing incorrect assessments. Finally, the metric’s reliance on English textual embeddings and specific datasets may inadvertently privilege certain languages and cultures.

\section*{Acknowledgments}
This research was supported in part by the U.S. National Science Foundation under Award No. 2418664. We thank Asteria Kaeberlein and Katherine Atwell for their helpful feedback. 

% \section*{Acknowledgments}

\bibliographystyle{acl_natbib}
% \bibliography{anthology,ranlp2025}

\appendix
\section{Kernel Density Plots}
\label{sec:density-plots}
\subsection{Plots for PHOENIX-14T}
\begin{figure}[htbp]
    \centering
    \includegraphics[width=\linewidth]{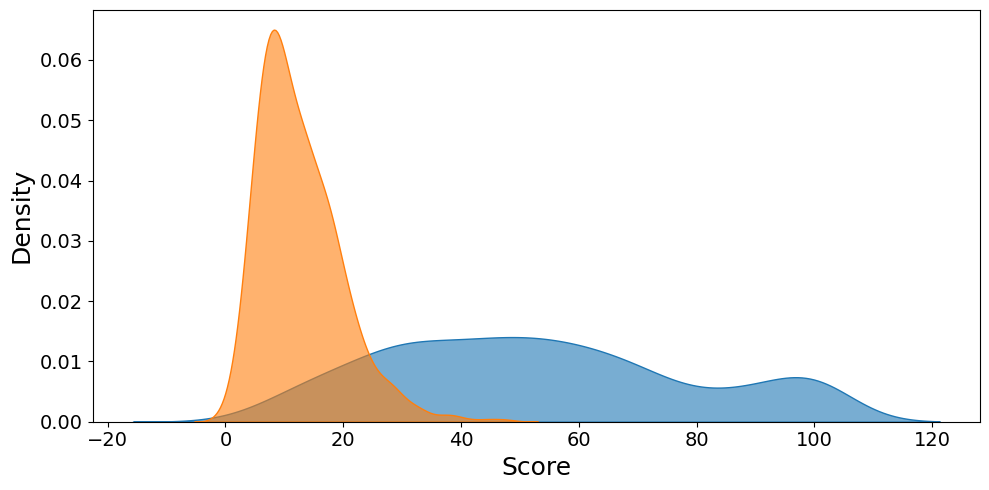}
    \includegraphics[width=\linewidth]{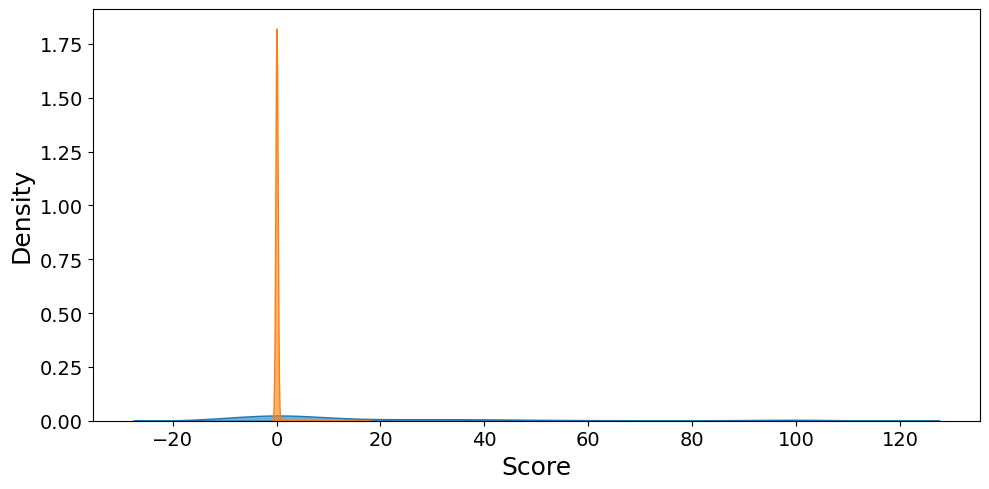}
    \caption{Kernel density plots for BLEU-1 (top) and BLEU-4 (bottom).}
    \label{fig:kde_app}
\end{figure}

Figure~\ref{fig:kde_app} shows KDEs of BLEU-1 (top) and BLEU-4 (bottom) scores on PHOENIX-14T. 

\subsection{Plots for CSL-Daily}
\begin{figure*}[ht]
    \centering
    \includegraphics[width=0.325\linewidth]{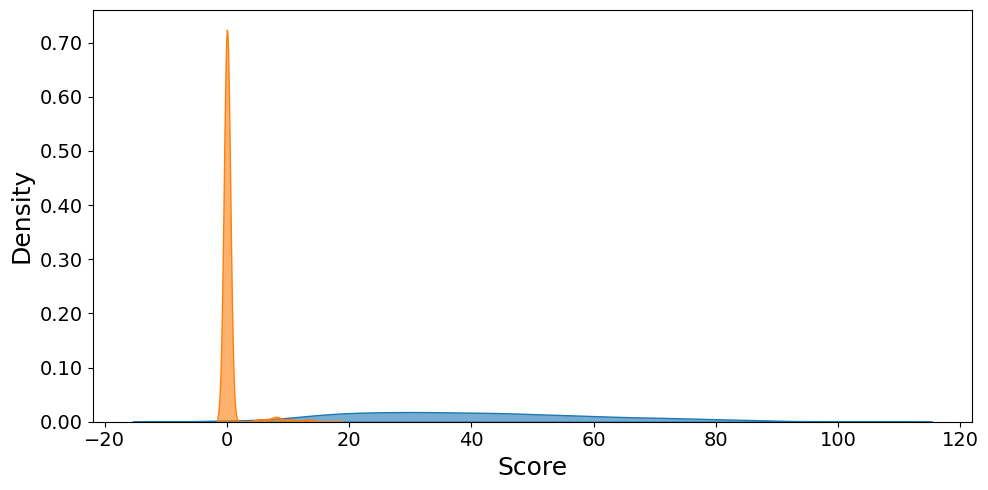}
    \includegraphics[width=0.325\linewidth]{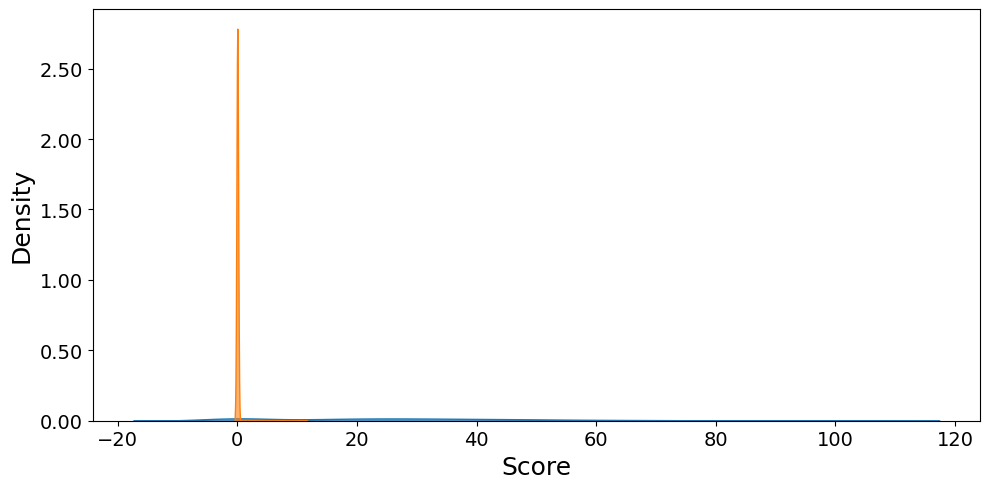}
    \includegraphics[width=0.325\linewidth]{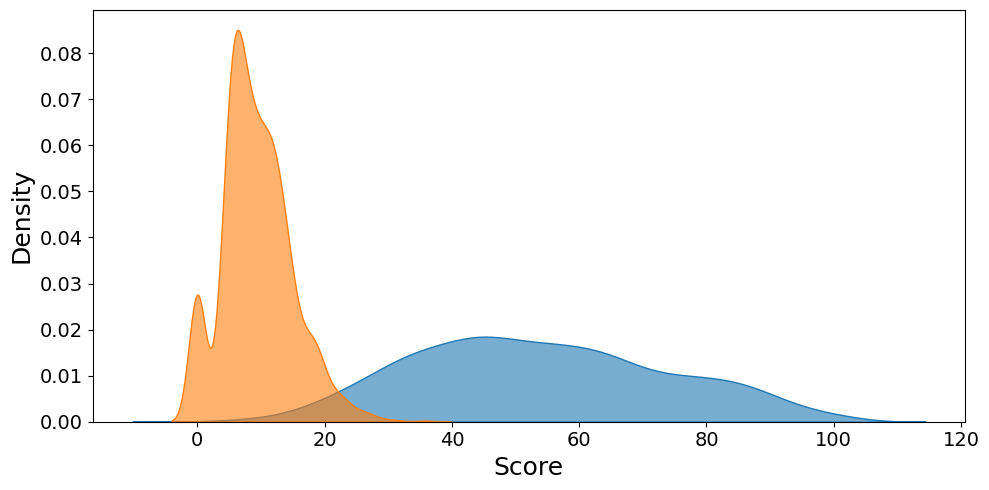}
    
    \includegraphics[width=0.325\linewidth]{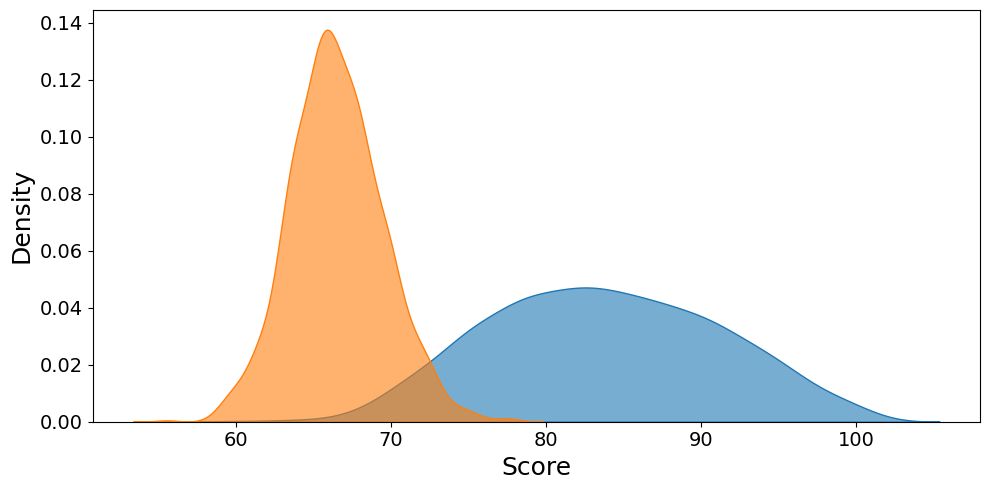}
    \includegraphics[width=0.325\linewidth]{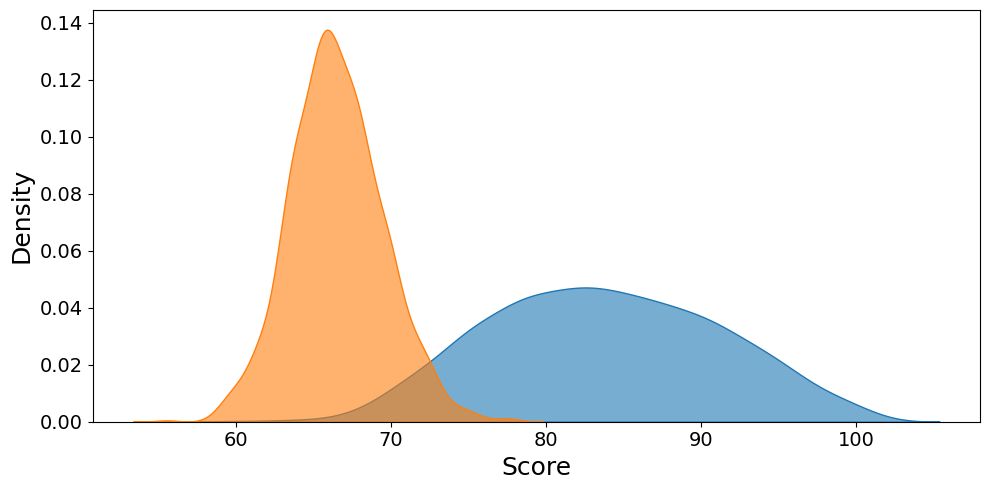}
    \includegraphics[width=0.325\linewidth]{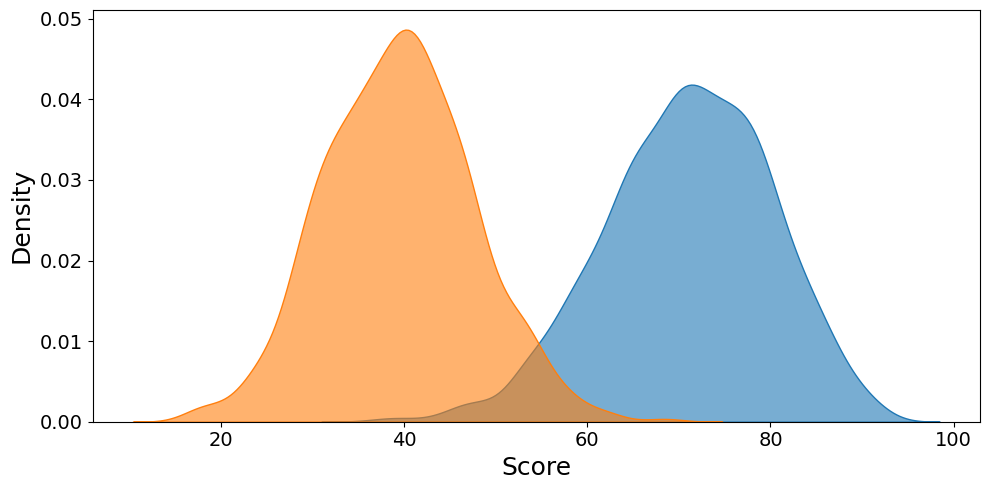}
    
    \caption{Kernel Density Plots for different metrics for CSL-Daily dataset. \textbf{Top row (left to right, rule-based metrics):} BLEU-2, BLEU-3, ROUGE. \textbf{Bottom row (left to right, embedding-based metrics):} BERTScore, BLEURT, SiLVERScore. SiLVERScore exhibits a clear separation between the two distributions, indicating a strong ability to distinguish aligned from misaligned pairs. }
    \label{fig:kdp_csl}
\end{figure*}

Figure~\ref{fig:kdp_csl} reports KDEs for six metrics on CSL-Daily. The rule-based metrics (BLEU-2/3, ROUGE) concentrate near lower scores with limited dynamic range, whereas the embedding-based metrics (BERTScore, BLEURT, SiLVERScore) yield broader, more informative distributions. SiLVERScore produces the clearest separation between the two distributions (aligned vs. misaligned pairs), suggesting stronger discrimination of semantic alignment.

\section{CSL-Daily results for Semantic test}
\begin{figure}[h]
    \centering
    \includegraphics[width=1\linewidth]{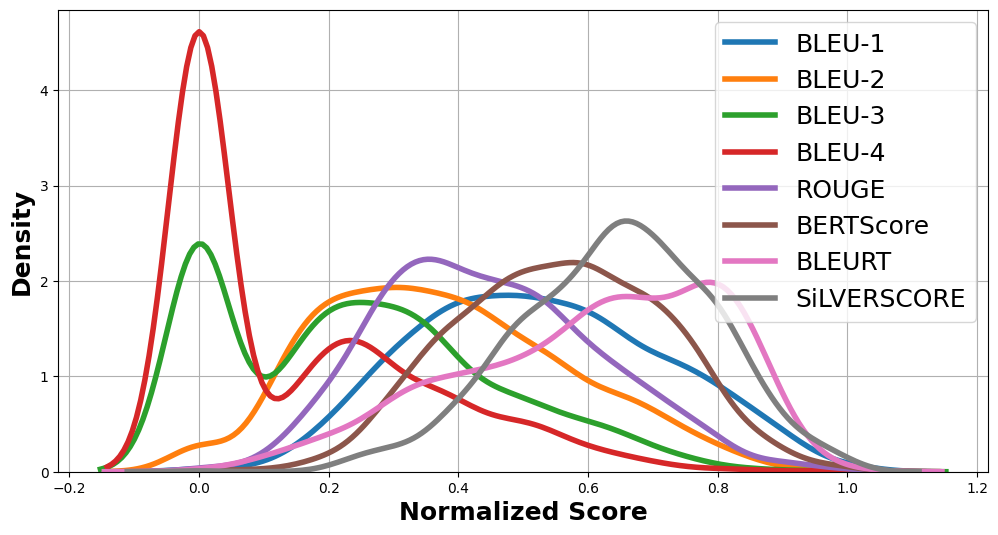}
    \caption{Kernel Density plots comparing the score distributions of different evaluation metrics when applied reordered hypotheses on CSL-Daily. SiLVERScore, BERTScore, and BLEURT show broader distributions and higher overlap, while rule-based metrics exhibit sharp peaks at lower scores. This indicates their sensitivity to surface-level word order changes.}
    \label{fig:kde_reordered_app}
\end{figure}

Figure~\ref{fig:kde_reordered_app} visualizes word order sensitivity by evaluating reordered hypotheses on CSL-Daily. Rule-based metrics collapse into sharp peaks at low scores, which reveals a high sensitivity to surface reordering. In contrast, embedding-based metrics (SiLVERScore, BERTScore, BLEURT) remain robust to paraphrastic reordering.
\label{sec:csl-daily-app}

\section{Heatmaps}
\begin{figure}[htbp]
    \centering
    \includegraphics[width=\linewidth]{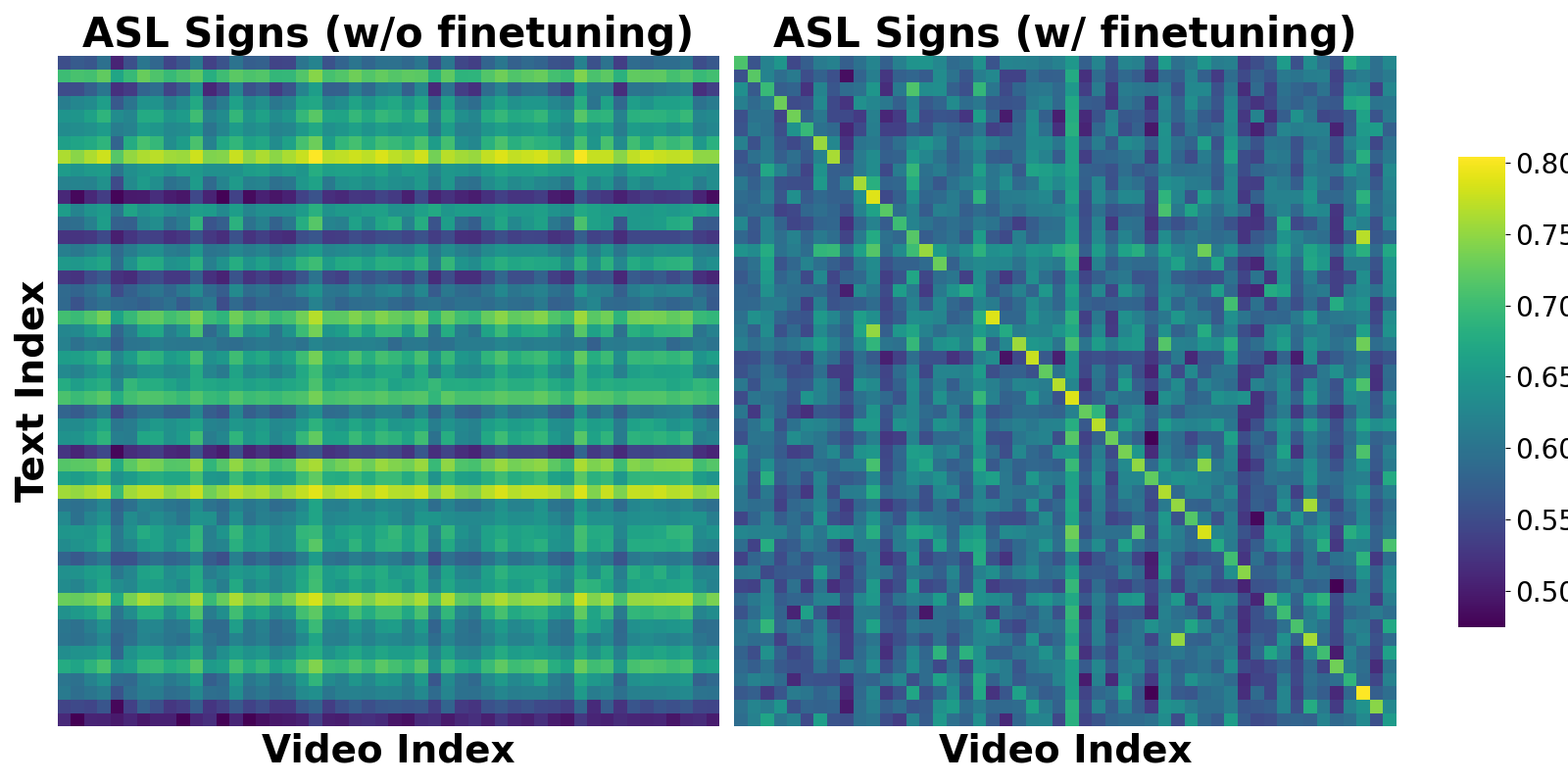}
    \caption{Heatmaps of SignCLIP embeddings cosine similarity scores for ASL Signs (token level).}
    \label{fig:heatmaps_app}
\end{figure}

Figure~\ref{fig:heatmaps_app} shows a heatmap of SignCLIP embedding cosine similarity scores for the ASL Signs dataset. A sharper diagonal pattern on the right indicates increased alignment between sign embeddings and their corresponding references.
\label{sec:heatmaps}

\begin{figure}[htbp]
    \centering
    \includegraphics[width=\linewidth]{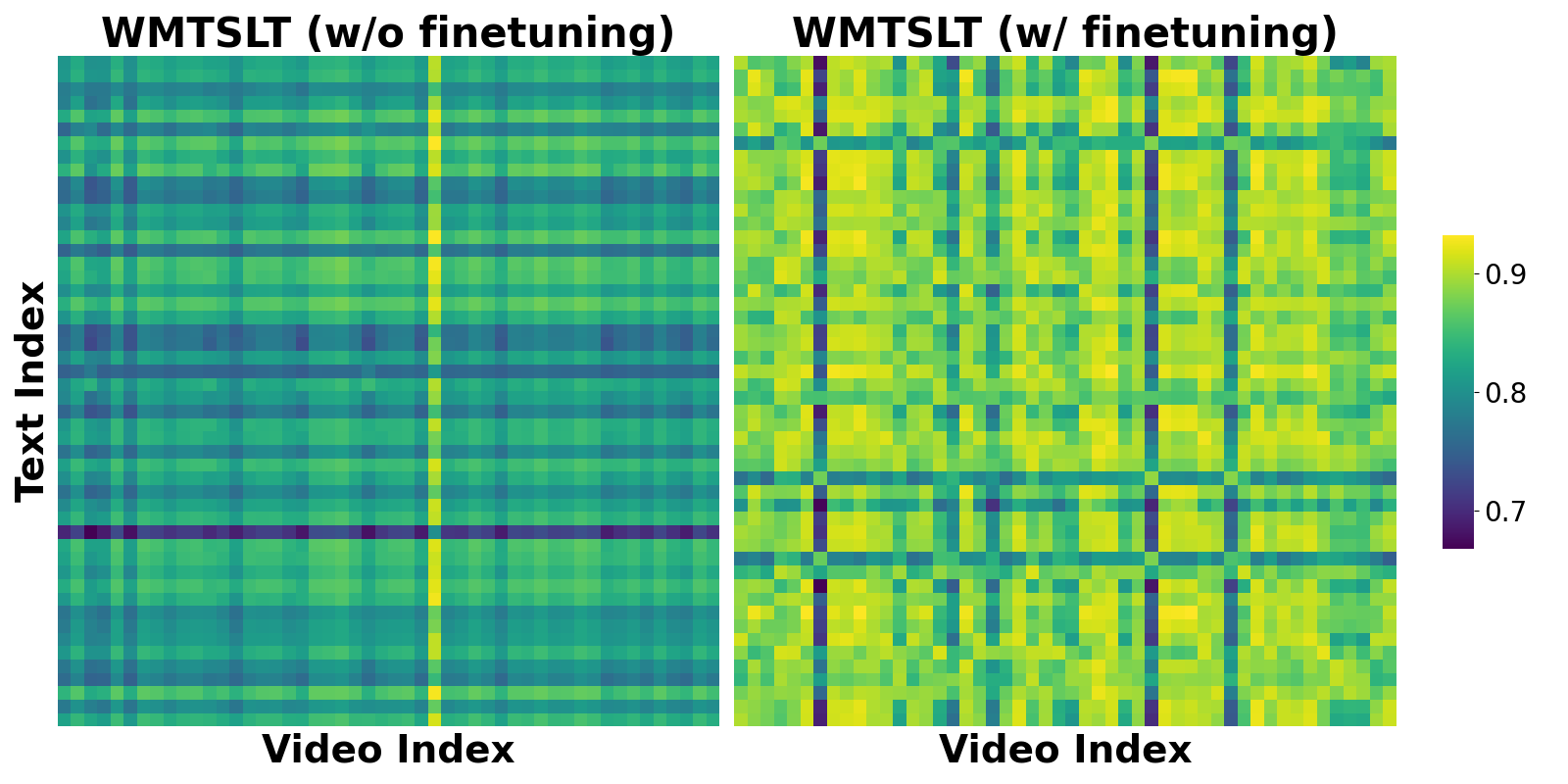}
    \caption{Heatmaps of SignCLIP embeddings cosine similarity scores for WMTSLT (sentence level).}
    \label{fig:heatmaps_app2}
\end{figure}

Figure~\ref{fig:heatmaps_app2} shows a heatmap of SignCLIP embedding cosine similarity scores for the WMTSLT dataset. After fine-tuning, the model appears to overfit, assigning high similarity scores to many pairs.

\section{Training Details}
\paragraph{Hardware and Compute} All training and inference computations were performed on an NVIDIA A100 GPU with 80GB of GPU memory. The experiments were conducted on a Linux-based server environment equipped with 8 CPU cores.
\paragraph{SignCLIP Fine-Tuning} Fine-tuning the SignCLIP model on American Sign Language (ASL) datasets took the longest (approximately 4 hours). Fine-tuning was conducted using a batch size of 256, a maximum length of 64, Adam optimizer with $\beta_1=0.9, \beta_2=0.98$, and a gradient clipping norm of 2.0. Training involved 1,000,000 total updates and a warm-up phase over the first 122 updates. Up to 25 epochs were run with 1000 steps for monitoring.
\paragraph{Sign Language Translation (MSKA) Inference} Inference with the MSKA model for sign language translation completed in under 10 minutes. Pre-trained weights from \citealp{Guan2024MultiStreamKA} were used without further fine-tuning. 

\section{Prosody Boxplots}
\begin{figure*}[h]
    \centering
    \includegraphics[width=1\linewidth]{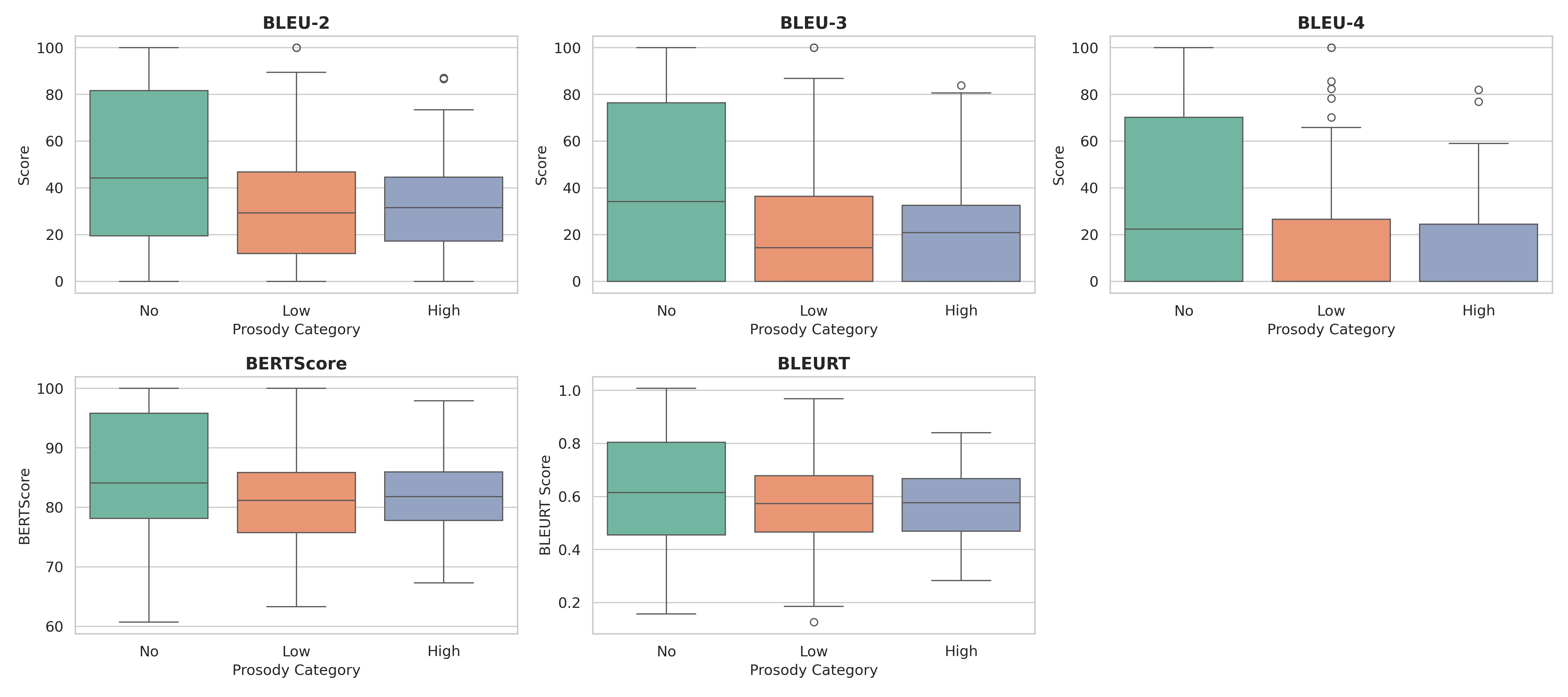}
    \caption{Box plots showing the distribution of BLEU-2, BLEU-3, BLEU-4, BERTScore, and BLEURT scores across three prosody intensity categories (No Intensity, Low Intensity, and High Intensity). Traditional back-translation metrics (BLEU) and embedding-based metrics (BERTScore and BLEURT) show a decline in scores with increasing prosody intensity}
    \label{fig:prosody_app}
\end{figure*}

Figure~\ref{fig:prosody_app} on the following page illustrates how various evaluation metrics distribute across sentences with different levels of prosodic intensity (No Intensity, Low Intensity, and High Intensity).
\label{sec:boxplots}

\section{Dataset Specifications}
Table~\ref{tab:dataset_specifications} summarizes the datasets used for our experiments in \S~\ref{sec:generalization_problem}, covering American Sign Language (ASL) and Swiss German Sign Language (DSGS).
\label{sec:dataset_details}
\begin{table*}[h!]
\centering
\begin{tabular}{lcccc}
\hline
\textbf{Dataset}      & \textbf{Language} & \textbf{Level}       & \textbf{\# of Samples (Train/Val/Test)} & \textbf{\# of Signers}        \\ \hline
\textbf{ASL Signs}    & ASL               & Token Level          & 85,031 / 4,723 / 4,723                     & 100+ Signers                            \\ 
\textbf{SemLex}       & ASL               & Token Level          & 51,029 / 18,025 / 15,514               & 119 deaf signers              \\ 
\textbf{ASL Citizen}  & ASL               & Token Level          & 40,154 / 10,304 / 32,941               & 52 deaf/hard-of-hearing       \\ 
\textbf{WMTSLT}       & DSGS              & Sentence Level       &  9172 / 470 / 494*                                     & 12 deaf signers               \\ \hline
\end{tabular}
\caption{Overview of the datasets used in our evaluations. For the WMTSLT dataset, the train/validation/test split was generated by the authors, as the original dataset provided by the challenge did not include a predefined test set.}
\label{tab:dataset_specifications}
\end{table*}

\section{Case Studies}
\subsection{PHOENIX-14T}
Table~\ref{tab:phoneix_case_study} presents three diverse examples from the PHOENIX-14T test set, including the original German text, its English translation, and each metric’s score. We choose these examples to highlight (1) how SiLVERScore handles complex weather-related descriptions, due to domain-aware fine-tuning on the contrastive learning model, (2) its robustness to high prosody, and (3) a short utterance and (4) negation, where it fails to capture the semantics.

\begin{table*}[h]
\centering
\resizebox{\textwidth}{!}{
\begin{tabular}{l p{5.5cm} p{5.5cm} ccccc}
\hline
\textbf{Case} 
  & \textbf{Reference (Human)} 
  & \textbf{System Output} 
  & \textbf{BLEU2} 
  & \textbf{ROUGE} 
  & \textbf{BERT} 
  & \textbf{BLEURT} 
  & \textbf{SiLVER}\\
\hline

\textbf{1: Complex} 
& \textit{regen und schnee lassen an den alpen in der nacht nach im norden und nordosten fallen hier und da schauer sonst ist das klar .}\newline
  \textit{\small (Rain and snow will fall in the Alps during the night, with showers here and there in the north and northeast, otherwise it will be clear.)} 
& \textit{die regenwolken lassen im laufe des tages nach am alpenrand schwächt sich ein wenig ab im norden und nordosten regnet es mitunter kräftig sonst zeigen sich auch die sterne .}\newline
  \textit{\small (The rain clouds will subside during the day and will weaken somewhat at the edge of the Alps. In the north and northeast it will rain heavily at times and the stars will also appear.)}
& 0.17 & 0.26 & 0.37 & 0.38 & 0.97 \\
 
\textbf{2: High prosody} 
& \textit{vom nordmeer zieht ein kräftiges tief heran und bringt uns ab den morgenstunden heftige schneefälle zum teil auch gefrierenden regen .}\newline
  \textit{\small (A strong low pressure system is moving in from the North Sea and will bring us heavy snowfalls and sometimes freezing rain from the morning onwards.)} 
& \textit{ein kräftiges tief über der nordsee sorgt ab morgen früh für teilweise kräftige schneefälle und gefrierenden regen .}\newline
  \textit{\small (A strong low over the North Sea will cause partly heavy snowfalls and freezing rain from tomorrow morning.)}
& 0.29 & 0.38 & 0.56 & 0.67 & 0.73 \\

\textbf{3: Short utterance (Failure)} 
& \textit{im westen ist es freundlich .}\newline
  \textit{\small (In the west it is pleasant.)} 
& \textit{im nordwesten freundlicher .}\newline
  \textit{\small (friendlier in the northwest.)}
& 0 & 0.35 & 0.54 & 0.41 & 0.69 \\

\textbf{4: Negation (Failure)} 
& \textit{heute nacht noch nicht so ganz da haben wir noch ein bisschen mit regen zu kämpfen an den küsten .}\newline
  \textit{\small (Not quite so much tonight as we still have a bit of rain to contend with on the coast.)} 
& \textit{heute nacht wird es hier und da noch etwas regnen an der nordseeküste .}\newline
  \textit{\small (There will be some rain here and there on the North Sea coast tonight.)}
& 0.11 & 0.31 & 0.37 & 0.59 & 0.43 \\

\hline
\end{tabular}
}
\caption{Case study with original German text and English translations for PHOENIX-14T dataset.}
\label{tab:phoneix_case_study}
\end{table*}

\subsection{CSL-Daily}
Table~\ref{tab:csl_case_study} shows four examples from the CSL-Daily dataset, which features Chinese Sign Language in various everyday contexts. We selected these cases to demonstrate (1) how SiLVERScore can handle more advanced sentences beyond casual daily conversation, (2) how it remains robust even when incorrect translations cause bottlenecks in proper evaluation for back-translation methods, and (3) a failure scenario in which limited context undermines its semantic alignment.

\begin{table*}[h]
\centering
\resizebox{\textwidth}{!}{%
\begin{CJK*}{UTF8}{gkai}
\begin{tabular}{l p{5.5cm} p{5.5cm} ccccc}
\hline
\textbf{Case} 
  & \textbf{Reference (Human)} 
  & \textbf{System Output} 
  & \textbf{BLEU-2} 
  & \textbf{ROUGE} 
  & \textbf{BERT} 
  & \textbf{BLEURT}
  & \textbf{SiLVER}\\
\hline

\textbf{1: Complex} 
& \textit{政府机关的改革，不但要精简多余的人员，更要减少重叠的机构。}\newline
  \textit{\small (The reform of government agencies should not only streamline redundant personnel, but also reduce overlapping institutions.)} 
& \textit{政客们在改革后,不但废除了改革的那些机会,也废除了改革的那些机会。}\newline
  \textit{\small (After the reforms, politicians not only abolished the opportunities for reform;)}
& 0.14 & 0.29 & 0.41 & 0.41 & 0.74\\

\textbf{2: Incomplete translation 1} 
& \textit{家里生活条件好了，奶奶还保持着勤俭节约的好习惯。}\newline
  \textit{\small (Although the family’s living conditions have improved, grandma still maintains the good habit of diligence and thrift.)} 
& \textit{奶奶老奶奶老是留着一副节约的好习惯。}\newline
  \textit{\small (Grandma always has a good habit of saving.)}
& 0.34 & 0.46 & 0.51 & 0.50 & 0.93\\
 
\textbf{3: Incomplete translation 2} 
& \textit{他的女儿每天晚上写作业写到很晚。}\newline
  \textit{\small (His daughter does her homework until very late every night.)} 
& \textit{女儿每天写作很晚,每天晚上。}\newline
  \textit{\small (My daughter writes very late every day, every night.)}
& 0.56 & 0.59 & 0.68 & 0.52 & 0.81\\

\textbf{4: Complex (Failure)} 
& \textit{为读者的家居生活带来不同凡响的流行资讯。}\newline
  \textit{\small (Bring extraordinary popular information to readers' home life.)} 
& \textit{为了学校环境好,我随身携带了不流行的新闻。}\newline
  \textit{\small (For the sake of a good school environment, I carried unpopular news with me.)}
& 0.13 & 0.29 & 0.37 & 0.31 & 0.49 \\

\hline
\end{tabular}
\end{CJK*}
}
\caption{Case study with original Chinese text and English translations for CSL-Daily dataset.}
\label{tab:csl_case_study}
\end{table*}

\subsection{Findings}
Across both datasets, SiLVERScore consistently outperforms back-translation based metrics and captures prosodic and lexical variations more effectively. However, it can fail in scenarios involving negations and when utterances are short. Future enhancements could incorporate discourse context and explicit modeling of negation or fine-grained sign features to make SiLVERScore even more sensitive to subtle meaning shifts.

\end{document}